\documentclass{article} 
\usepackage{nips11submit_e,times}

\usepackage{latexsym}
\usepackage{amsmath}
\usepackage{url}
\usepackage{graphicx}

\usepackage{wrapfig,comment}
\usepackage{bbm}

\usepackage{lastpage,graphics,graphicx}
\usepackage{amsmath,amssymb,amsfonts}
\usepackage{url}

\DeclareMathOperator*{\softmax}{softmax}

\usepackage{amsthm}

\usepackage{enumitem}
   {
      \begin{enumerate}[nolistsep,label=\Alph*.,ref=\Alph{enumi},
        leftmargin=0pt,labelsep=15pt,align=left,
        labelwidth=0.1cm,itemindent=1.0cm]}
  {\end{enumerate}}
\usepackage{enumitem}
   {
      \begin{enumerate}[nolistsep,label=\Alph*.,ref=\Alph{enumi},
        leftmargin=0pt,labelsep=15pt,align=left,
        labelwidth=0.1cm,itemindent=1.0cm]}
  {\end{enumerate}}
   {
      \begin{enumerate}[nolistsep,label=\Alph*.,ref=\Alph{enumi},
        leftmargin=0pt,labelsep=15pt,align=left,
        labelwidth=1.5cm,itemindent=2.0cm]}
  {\end{enumerate}}

\title{Zero-Shot Learning Through Cross-Modal Transfer} 

\author{\hspace{-0.3cm}Richard Socher,  Milind Ganjoo, Hamsa Sridhar, Osbert Bastani, Christopher D. Manning, Andrew Y. Ng\\
Computer Science Department, Stanford University, Stanford, CA 94305, USA\\
{\tt \small richard@socher.org, \{mganjoo, hsridhar, obastani, manning, ang\}@stanford.edu}\\
}

\nipsfinalcopy 

\begin{document}

\maketitle

\begin{abstract}
This work introduces a model that can recognize objects in images even if no training data is available for the objects. The only necessary knowledge about the unseen categories comes from unsupervised large text corpora. In our zero-shot framework distributional information in language can be seen as spanning a semantic basis for understanding what objects look like.
Most previous zero-shot learning models can only differentiate between unseen classes. In contrast, our model can both obtain state of the art performance on classes that have thousands of training images and obtain reasonable performance on unseen classes. This is achieved by first using outlier detection in the semantic space and then two separate recognition models.
Furthermore, our model does not require any manually defined semantic features for either words or images. 
\end{abstract}

\section{Introduction}
The ability to classify instances of an unseen visual class, called zero-shot learning, is useful in many situations. There are many species, products or activities without labeled data and new visual categories, such as the latest gadgets or car models are introduced frequently. 
In this work, we show how to make use of the vast amount of knowledge about the visual world available in natural language to classify unseen objects. 
We attempt to model people's ability to identify unseen objects even if the only knowledge about that object came from reading about it. 
For instance, after reading the description of 
\emph{a two-wheeled self-balancing electric vehicle, controlled by a stick, with which you can move around while standing on top of it}, 
many would be able to identify a 
\emph{Segway}, possibly after being briefly perplexed because the new object looks different to any previously observed object class. 

We introduce a zero-shot model that can predict both seen and unseen classes. For instance, without ever seeing a cat image, it can determine whether an image shows a cat or a known category from the training set such as a dog or a horse. The model is based on two main ideas. 

First, images are mapped into a semantic space of words that is learned by a neural network model \cite{Huang2012}. Word vectors capture distributional similarities from a large, unsupervised text corpus. By learning an image mapping into this space, the word vectors get implicitly grounded by the visual modality, allowing us to give prototypical instances for various words.

\begin{figure}[t]
\centering
\includegraphics[width=1.03\columnwidth]{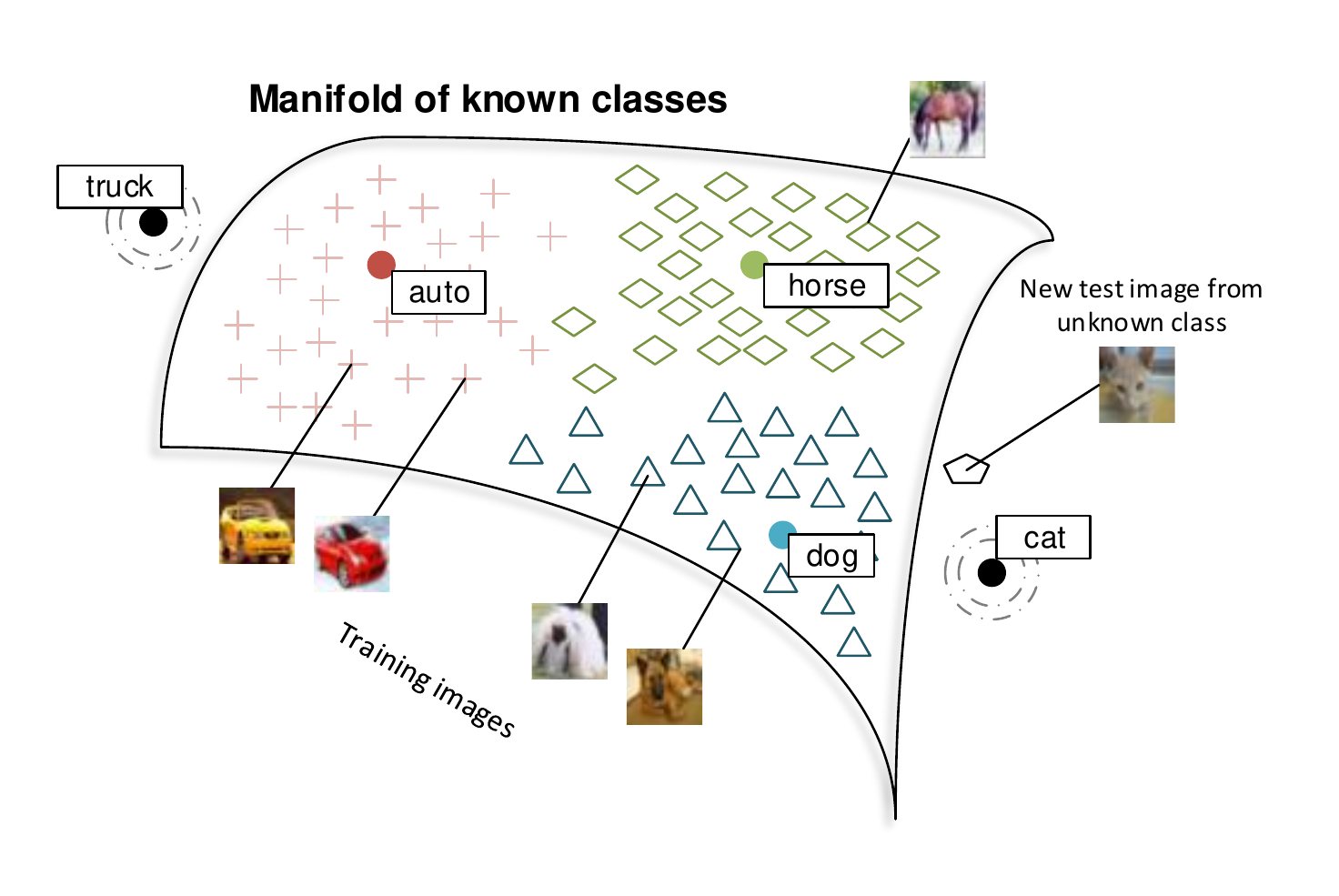}\\
\caption{Overview of our multi-modal zero-shot model. We first map each new testing image into a lower dimensional semantic space. Then, we use outlier detection to determine whether it is on the manifold of seen images. If the image is not on the manifold, we determine its class with the help of unsupervised semantic word vectors. In this example, the unseen classes are truck and cat.}
\label{fig1}
\end{figure}

Second, because classifiers prefer to assign test images into classes for which they have seen training examples, the model incorporates an outlier detection probability which determines whether a new image is on the manifold of known categories. If the image is of a known category, a standard classifier can be used. Otherwise, images are assigned to a class based on the likelihood of being an unseen category. The probability of being an outlier or a known category is integrated into our probabilistic model. 
The model is illustrated in Fig \ref{fig1}.

Unlike previous work on zero-shot learning which can only predict intermediate features or differentiate between various zero-shot classes \cite{zeroShotSemanticCodes}, our joint model can achieve both state of the art accuracy on known classes as well as reasonable performance on unseen classes.
Furthermore, compared to related work in knowledge transfer \cite{attributeTransferLampert,crossCatProp} we do not require manually defined semantic or visual attributes for the zero-shot classes. Our language feature representations are learned from unsupervised and unaligned corpora.

We first briefly describe a selection of related work, followed by the model description and experiments on CIFAR10.

\section{Related Work}
We briefly outline connections and differences to five related lines of research. Due to space constraints, we cannot do justice to the complete literature.

\textbf{Zero-Shot Learning. } 
The work most similar to ours is that by Palatucci et al. \cite{zeroShotSemanticCodes}. They map fMRI scans of people thinking about certain words into a space of manually designed features and then classify using these features. They are able to predict semantic features even for words for which they have not seen scans and experiment with differentiating between several zero-shot classes. However, the do not classify new test instances into both seen and unseen classes. We extend their approach to allow for this setup using outlier detection. 

Larochelle et al. \cite{zeroData} describe the unseen zero-shot classes by a ``canonical'' example or use ground truth human labeling of attributes.

\textbf{One-Shot Learning}
One-shot learning \cite{FFOneShot,oneShotLake} seeks to learn a visual object class by using very few training examples. This is usually achieved by either sharing of feature representations \cite{BartOneShot}, model parameters \cite{finkOneShot} or via similar context \cite{hoiemOneShot}. A recent related work on one-shot learning is that of Salakhutdinov et al. \cite{rusOneShot}. Similar to their work, our model is based on using deep learning techniques to learn low-level image features followed by a probabilistic model to transfer knowledge. However, our work is able to classify object categories without any training data due to the cross-modal knowledge transfer from natural language and at the same time obtain high performance on classes with many training examples.

\textbf{Knowledge and Visual Attribute Transfer. } 
Lambert et al.\ and Farhadi et al.\ \cite{attributeTransferLampert,Farhadi} were two of the first to use well-designed visual attributes of unseen classes to classify them. This is different to our setting since we only have distributional features of words learned from unsupervised, non-parallel corpora and can classify between categories that have thousands or zero training images.
Qi et al. \cite{crossCatProp} learn when to transfer knowledge from one category to another for each instance. 


\textbf{Domain Adaptation. } 
Domain adaptation is useful in situations in which there is a lot of training data in one domain but little to none in another. For instance, in sentiment analysis one could train a classifier for movie reviews and then adapt from that domain to book reviews \cite{blitzerDA,glorotDA}. While related, this line of work is different since there is data for each \emph{class} but the features may differ between domains.

\textbf{Multimodal Embeddings. }
Multimodal embeddings relate information from multiple sources such as sound and video \cite{ngiamMM} or images and text. Socher et al. \cite{SocherFeiFeiCVPR2010} project words and image regions into a common space using kernelized canonical correlation analysis to obtain state of the art performance in annotation and segmentation. Similar to our work, they use unsupervised large text corpora to learn semantic word representations. Their model does require a small amount of training data however for each class. Among other recent work is that by Srivastava and Salakhutdinov \cite{rusMM} who developed multimodal Deep Boltzmann Machines. Similar to their work, we use techniques from the broad field of deep learning to represent images and words. 

Some work has been done on multimodal distributional methods \cite{fengDistSem,mihaImTxt}. Most recently, Bruni et al. \cite{bruni2012} worked on perceptually grounding word meaning and showed that joint models are better able to predict the color of concrete objects.

\section{Word and Image Representations}
We begin the description of the full framework with the feature representations of words and images.
%
Distributional approaches are very common for capturing semantic similarity between words. In these approaches, words are represented as vectors of distributional characteristics -- most often their co-occurrences with words in context \cite{PadoAndLapata2007,Erk:2008,BaroniAndLenci2010,TurneyAndPantel2010}. These representations have proven very effective in natural language processing tasks such as sense disambiguation \cite{Schuetze1998}, thesaurus extraction \cite{Lin1998,Curran2004} and cognitive modeling \cite{LandauerAndDumais1997}. 

We initialize all word vectors with pre-trained $50$-dimensional word vectors from the unsupervised model of Huang et al. \cite{Huang2012}. Using free Wikipedia text, their model learns word vectors by predicting how likely it is for each word to occur in its context. Their model uses both local context in the window around each word and global document context.
Similar to other local co-occurrence based vector space models, the resulting word vectors capture distributional syntactic and semantic information.
For further details and evaluations of these embeddings, see \cite{Bengio2003,collobert2008:deep}.  

We use the unsupervised method of Coates et al. \cite{AdamKmeans} to extract $F$ image features from raw pixels in an unsupervised fashion. Each image is henceforth represented by a vector $x\in \mathbb{R}^F$. 

\section{Projecting Images into Semantic Word Spaces}
In order to learn semantic relationships and class membership of images we project the image feature vectors into the $50$-dimensional word space. 
During training and testing, we consider a set of classes $Y$. Some of the classes $y$ in this set will have available training data, others will be zero-shot classes without any training data. We define the former as the seen classes $Y_s$ and the latter as the unseen classes $Y_u$. 
Let $W=W_s \cup W_u$ be the set of word vectors capturing distributional information for both seen and unseen visual classes, respectively. 

All training images $x^{(i)} \in X_y$ of a seen class $y \in Y_s$ are mapped to the word vector $w_y$ corresponding to the class name. To train this mapping, we minimize the following objective function with respect to the matrix $\theta \in \mathbb{R}^{50\times F}$:
\begin{equation}
J(\theta) = \sum_{y\in Y_s} \sum_{x^{(i)} \in X_y} \| w_y - \theta x^{(i)}  \|^2.
\label{eq:map}
\end{equation}

By projecting images into the word vector space, we implicitly extend the word semantics with a visual grounding, allowing us to query the space, for instance for prototypical visual instances of a word or the average color of concrete nouns. 

\begin{figure}[t]
\vspace{-1.2cm}
\centering
\includegraphics[width=1.03\columnwidth]{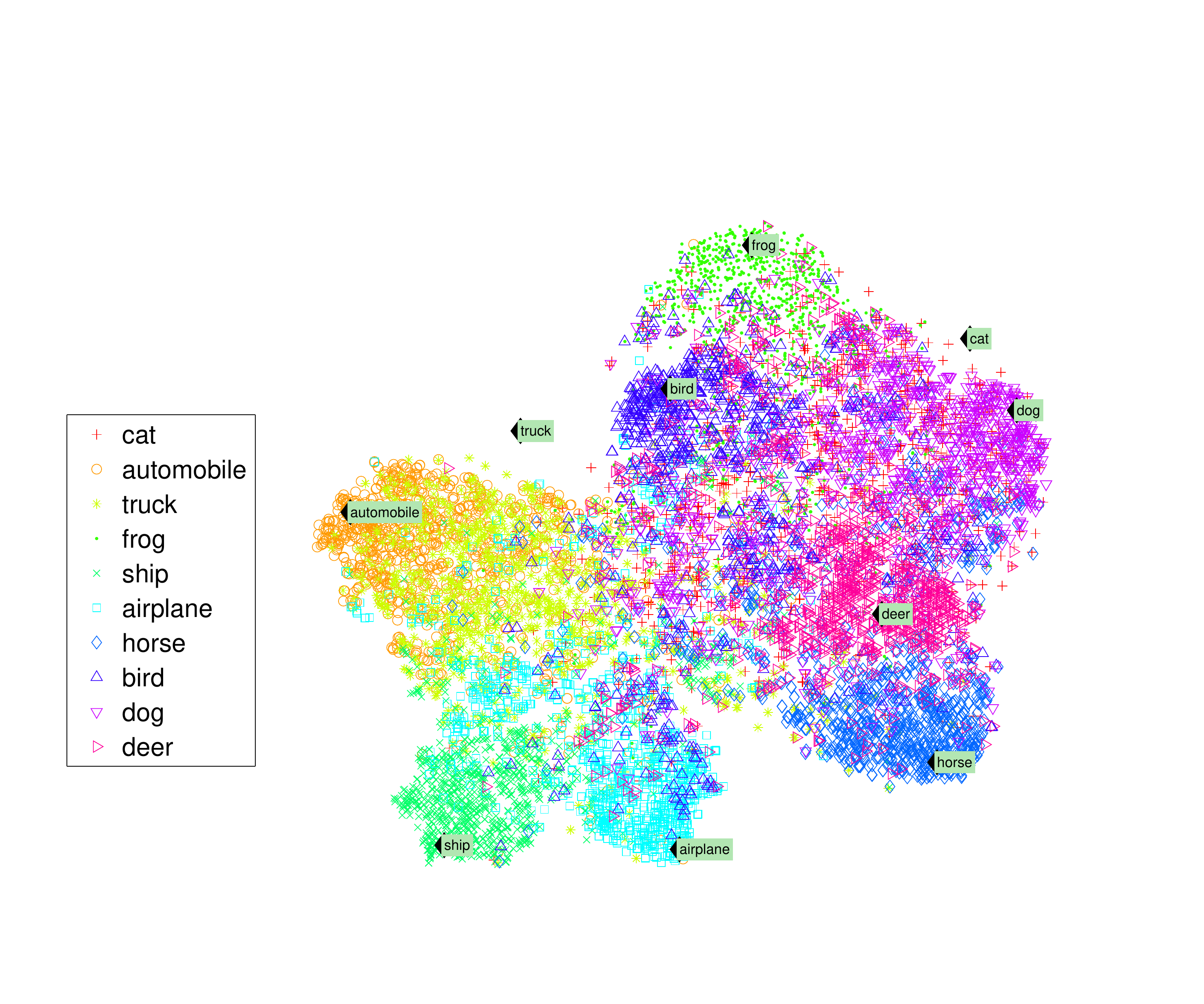}\\
\vspace{-1.0cm}
\caption{T-SNE visualization of the semantic word space. Word vector locations are highlighted and mapped image locations are shown both for images for which this mapping has been trained and unseen images.  The unseen classes are cat and truck.}
\label{fig:mapping}
\end{figure}

Fig. \ref{fig:mapping} shows a visualization of the 50-dimensional semantic space with word vectors and images of both seen and unseen classes. The unseen classes are cat and truck. The mapping from 50 to 2 dimensions was done with t-SNE \cite{tsne}. We can observe that most classes are tightly clustered around their corresponding word vector while the zero-shot classes (cat and truck for this mapping) do not have close-by vectors. However, the images of the two zero-shot classes are close to semantically similar classes. For instance, the cat testing images are mapped most closely to dog, and horse and are all very far away from car or ship. This motivated the idea for first finding outliers and then classifying them to the zero-shot word vectors.

Now that we have covered the representations for words and images as well as the image to word space mapping we can describe the probabilistic model for joint zero-shot learning and standard image classification.

\section{Zero-Shot Learning Model}
In this section we first give an overview of our model and then describe each of its components.
In general, we want to predict $p(y|x)$, the conditional probability for both seen and unseen classes $y\in Y_s\cup Y_u$ given an image $x$.
Because standard classifiers will never predict a class that has no training examples, we introduce a binary \emph{visibility} random variable which indicates whether an image is in a seen or unseen class $V\in \{s,u\}$. Let $X_s$ be the set of all feature vectors for training images of seen classes. 

We predict the class $y$ for a new input image $x$ via:
\begin{equation}
p(y|x,X_s,W,\theta) = \sum_{V\in \{s,u\}} P(y|V,x,X_s,W,\theta)P(V|x,X_s,W,\theta).
\label{eq:main}
\end{equation}
Next, we will describe each factor in Eq. \ref{eq:main}. 

The term $P(V=u|x,X_s,W,\theta)$ is the probability of an image being in an unseen class. It can be computed by thresholding an outlier detection score. This score is computed on the manifold of training images that were mapped to the semantic word space. We use a threshold on the marginal of each point under a mixture of Gaussians. 
The mapped points of seen classes are used to obtain this marginal:
$P(x|X_s,W_s,\theta) = \sum_{y\in  Y_s} P(x|y)P(y)= \sum_{y\in Y_s} \mathcal{N}(\theta x|w_y, \Sigma_y)P(y)$.  The Gaussian of each class is parameterized by the corresponding semantic word vector $w_y$ for its mean and a covariance matrix $\Sigma_y$ that is estimated from all the mapped training points with that label. We restrict the Gaussians to be isometric to prevent overfitting.

For a new image $x$, the outlier detector then becomes the indicator function that is $1$ if the marginal probability is below a certain threshold $T$:
\begin{equation}
P(V=u|x,X_s,W,\theta) := \mathbbm{1}\{P(x|X_s,W_s,\theta) < T\}
\label{eq:outlier}
\end{equation}
We provide an experimental analysis for various thresholds $T$ below. 

In the case where $V=s$, i.e. the point is considered to be of a known class, we can use any classifier for obtaining $P(y|V=s,x,X_s)$. We use a $\softmax$ classifier on the original $F$-dimensional features. For the zero-shot case where $V=u$ we assume an isometric Gaussian distribution around each of the zero-shot semantic word vectors. 

An alternative would be to use the method of Kriegel et al. \cite{outlierProb} to obtain an outlier probability for each testing point and then use the weighted combination of classifiers for both seen and unseen classes.

\section{Experiments}
We run most of our experiments on the CIFAR10 dataset. The dataset has 10 classes, each with 5000 $32\times 32\times 3$ RGB images. 
We use the unsupervised feature extraction method of Coates and Ng \cite{AdamKmeans} to obtain a 12,800-dimensional feature vector for each image.
In the following experiments, we omit the training images of 2 classes for the zero-shot analysis. 

\subsection{Zero-Shot Classes Only}
In this section we compare classification between only two zero-shot classes.
We observe that if there is no seen class that is remotely similar to the zero-shot classes, the performance is close to random. In other words, if the two zero-shot classes are the most similar classes and the seen classes do not properly span the subspace of the zero-shot classes then performance is poor. For instance, when \emph{cat} and \emph{dog} are taken out from training, the resulting zero-shot classification does not work well because none of the other 8 categories is similar enough to learn a good feature mapping. On the other hand, if \emph{cat} and \emph{truck} are taken out, then the cat vectors can be mapped to the word space thanks to transfer from \emph{dogs} and \emph{trucks} can be mapped thanks to \emph{car}, so the performance is very high.

\begin{figure}[t]
\vspace{-1.2cm}
\centering
\includegraphics[width=1.03\columnwidth]{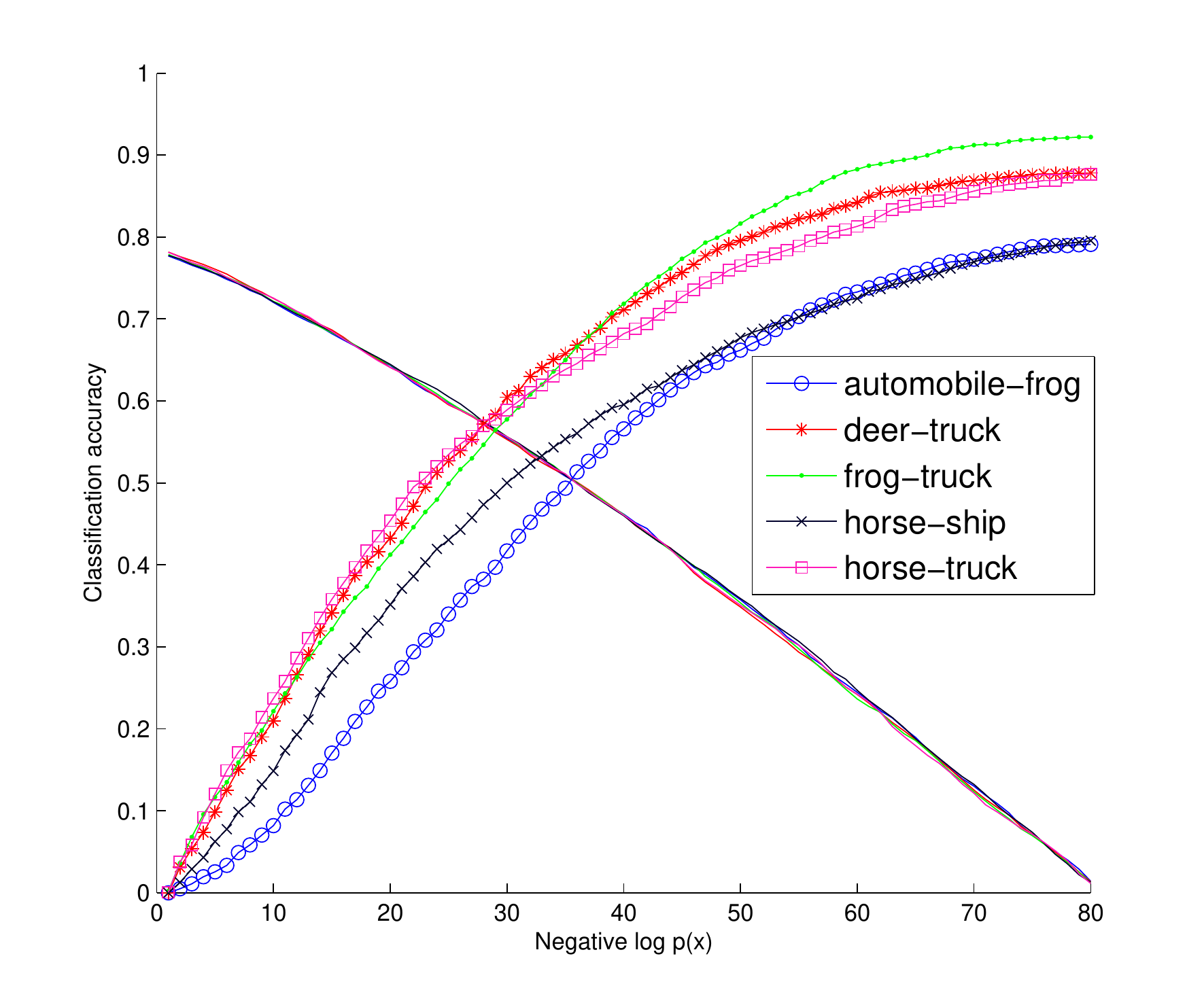}\\
\vspace{-0.5cm}
\caption{Visualization of the accuracy of the seen classes (lines from the top left to bottom right) and pairs of zero-shot classes (lines from bottom left to top right) at different thresholds of the negative log-likelihood of each mapped test image vector.}
\label{fig:main}
\end{figure}

Fig. \ref{fig:main} shows the performance at various cutoffs for the outlier detection. The cutoff is defined on the negative log-likelihood of the marginal of each point in the outlier detection. 
We can observe that when classifying images of unseen classes into only zero-shot classes (right side of the figure), we can differentiate images with an accuracy of above 80\%.

\subsection{Zero-Shot and Seen Classes}
In Fig. \ref{fig:main} we can observe that depending on the threshold that splits images into seen or unseen classes at test time we can obtain accuracies of trained classes of approximately 80\%. At 70\% accuracy, unseen classes can be classified with accuracies of between 30\% to 15\%. Random chance is 10\%.

\section{Conclusion}
We introduced a novel model for joint standard and zero-shot classification based on deep learned word and image representations. The two key ideas are that (i) using semantic word vector representations can help to transfer knowledge between categories even when these representations are learned in an unsupervised way and (ii) that our Bayesian framework that first differentiates outliers from points on the projected semantic manifold can help to combine both zero-shot and seen classification into one framework. If the task was only to differentiate between various zero-shot classes we could obtain accuracies of up to 90\% with a fully unsupervised model.


\small{
\bibliographystyle{plain}
\bibliography{bib}
}

\end{document}